\documentclass{article}

\usepackage{PRIMEarxiv}

\usepackage[utf8]{inputenc} 
\usepackage[T1]{fontenc}    
\usepackage{hyperref}       
\usepackage{url}            
\usepackage{booktabs}       
\usepackage{amsfonts}       
\usepackage{nicefrac}       
\usepackage{microtype}      
\usepackage{lipsum}
\usepackage{fancyhdr}       
\usepackage{graphicx}       
\graphicspath{{media/}}     

\usepackage{tabularx}
\usepackage[outdir=./]{epstopdf}
\usepackage{graphics}
\usepackage{verbatim}

\usepackage{multirow}
\usepackage{multicol}
\usepackage[table,xcdraw]{xcolor}
\usepackage{boldline}

\usepackage{cuted}
\usepackage{capt-of}

\usepackage{amsmath}

\title{Video Deblurring with Deconvolution and Aggregation Networks}

\author{
  Giyong Choi, HyunWook Park \\
  School of Electrical Engineering \\
  KAIST \\
  Daejeon \\
  \texttt{\{gychoi92, hwpark\}@kaist.ac.kr} \\
}

\begin{document}
\maketitle

\begin{abstract}
In contrast to single-image deblurring, video deblurring has the advantage that neighbor frames can be utilized to deblur a target frame. However, existing video deblurring algorithms often fail to properly employ the neighbor frames, resulting in sub-optimal performance. In this paper, we propose a deconvolution and aggregation network (DAN) for video deblurring that utilizes the information of neighbor frames well. In DAN, both deconvolution and aggregation strategies are achieved through three sub-networks: the preprocessing network (PPN) and the alignment-based deconvolution network (ABDN) for the deconvolution scheme; the frame aggregation network (FAN) for the aggregation scheme. In the deconvolution part, blurry inputs are first preprocessed by the PPN with non-local operations. Then, the output frames from the PPN are deblurred by the ABDN based on the frame alignment. In the FAN, these deblurred frames from the deconvolution part are combined into a latent frame according to reliability maps which infer pixel-wise sharpness. The proper combination of three sub-networks can achieve favorable performance on video deblurring by using the neighbor frames suitably. In experiments, the proposed DAN was demonstrated to be superior to existing state-of-the-art methods through both quantitative and qualitative evaluations on the public datasets.
\end{abstract}


\section{Introduction}

Videos, especially those obtained from hand-held devices, such as action cameras and smartphones, can be easily degraded by motion blurs caused by camera shake or object motion. Motions with a relatively long exposure time often lead to motion blurs on frames because the captured frame represents aggregate of multiple instants of moving objects during the exposure time. These motion blurs result in poor video quality and troubles for other major computer vision tasks, such as tracking \cite{jin2005visual, mei2008modeling} and object segmentation. Therefore, there has been an increasing demand for the removal motion blurs from videos.

The easiest way to reduce motion blurs in a video is to apply image deblurring techniques to each frame of video. Recently, many successful single-image deblurring algorithms have been proposed \cite{xu2017motion,kupyn2018deblurgan,tao2018scale}. A deblurred video can be obtained by simply using these methods. However, the image deblurring algorithms for each frame may fail to maintain temporal coherence through successive frames.

In contrast to image deblurring, video deblurring has the advantage that information of neighbor frames can be utilized when a target frame is deblurred. Because both camera and object motions are time dependent, some objects that appear blurry in a certain frame may appear sharp in adjacent frames. Therefore, a blurry frame can be deblurred efficiently by borrowing sharp pixels from its neighbor frames, and a better deblurring results can be obtained in comparison to using a single frame.

To make the best use of the information of neighbor frames, we need to examine which pixels in neighbor frames corresponding to the blurry pixels in the current frame and how sharp the pixels in the neighbor frames are. Therefore, even if an object in the current frame is found in its neighbor frames, its pixels should be exploited in accordance with the sharpness. To this end, Cho et al. \cite{cho2012video} searched patches similar to a current patch from its adjacent frames. Then, they measured luckiness, which inferred how sharp the patch was, of patches from neighbor frames and combined them using their own luckiness as weights to deblur the current patch. The neighbor patches were combined in the image domain in \cite{cho2012video}, whereas in \cite{delbracio2015hand}, they were combined according to the Fourier spectrum magnitude as weights in the Fourier domain. These methods can be regarded as the aggregation-based video deblurring approach. However, the performance improvement enabled by these aggregation-based methods may be insufficient because they employ neighbor frames without any restoration. When all neighbor frames are blur, the target frame cannot be deblurred well.

In a deep-learning-based video deblurring approach \cite{su2017deep}, multiple consecutive frames are stacked for an input of the network to utilize the information of neighbor frames. The network then extracts features from both a central frame and its neighbor frames to deblur the central frame, which results in better deblurring results than when only a single central frame is used. In \cite{su2017deep}, homography or optical flow estimation is adopted to align input frames. However, because the blurry pixels interfere with the homography or optical flow estimation between adjacent frames, it causes other undesired artifacts due to misalignment. Meanwhile, Wang et al. \cite{41wang2019edvr} employed deformable convolution \cite{dai2017deformable} to align input frames. In \cite{41wang2019edvr}, the neighbor frames were warped to the central frame well by using the deformable convolution. However, it is difficult to determine unreliable regions such as occlusion in the warped neighbor frames in the deformable-convolution-based frame alignment because the offset values of the deformable convolution are too messy to analyze.

To align input frames well, they should be sharp enough to estimate accurate object motion. On the contrary, to deblur the central frame effectively, its neighbor frames should be aligned with it properly. Thus, deblurring and alignment can be considered as a combined problem. To solve this combined problem, conventional deconvolution-based deblurring algorithms \cite{zhang2014multi,pan2017simultaneous} simultaneously handle the alignment and deblurring steps. These deconvolution-based methods formulate blur models which represent the relationship between blur frames and latent frames, and obtain deblurred frames by inversely solving the blur model. However, they require complex procedures to minimize the cost function and strongly depend on the energy model.

\begin{figure*} [!t]
\centering
\includegraphics[width=0.9\textwidth]{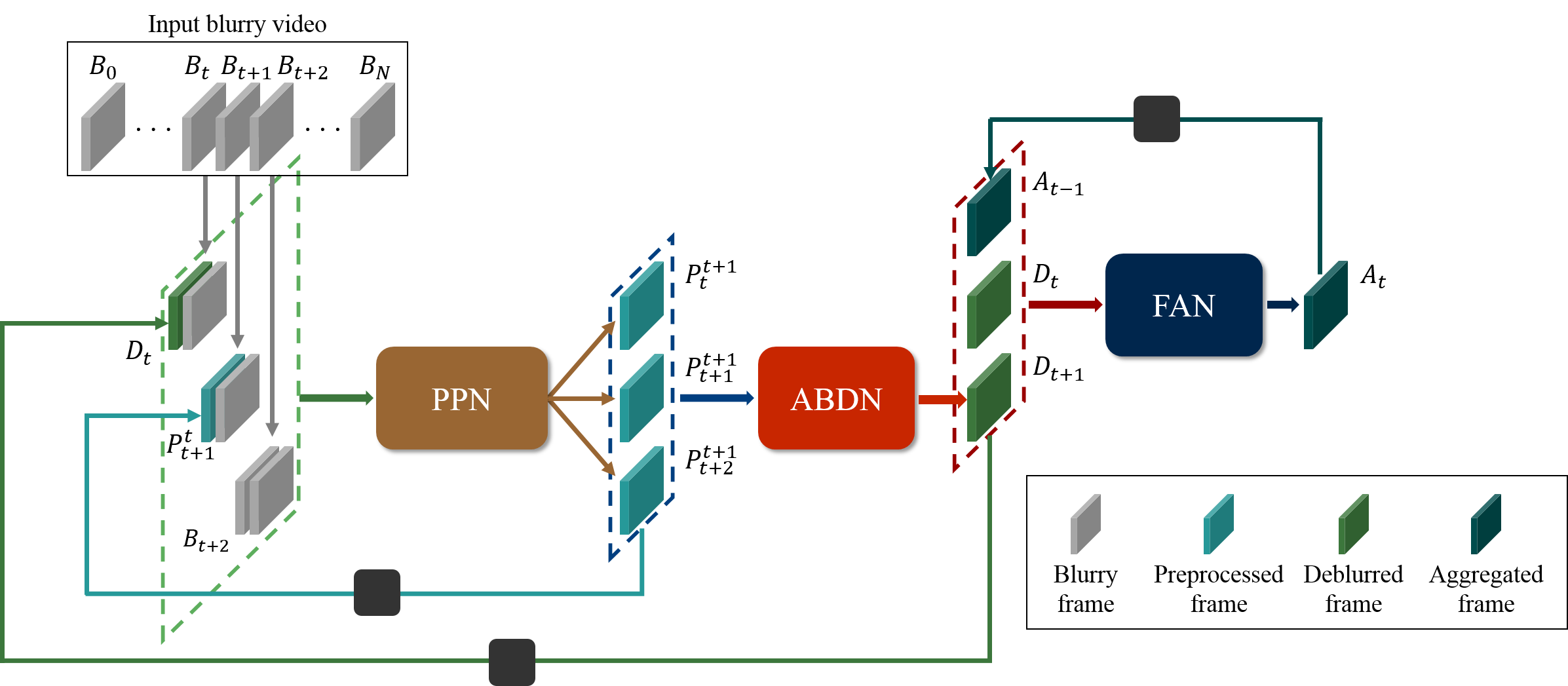}
\caption{Overall architecture of the deconvolution and aggregation network (DAN). Each black square indicates the delay of a single time step.}
\label{fig:overall}
\end{figure*}

Another group of methods \cite{hyun2017online,nah2019recurrent,wieschollek2017learning,zhou2019spatio} adopted recurrent neural networks (RNNs) to utilize information of the neighbor frames. Because the RNNs of these methods receive the previously restored frames, rather than a blurry one, as well as the current frame, the networks can utilize most of the information of neighbor frames. However, it is hard for a restored single frame to contain information of all previous restored frames. Besides, because the frame alignment is performed with the restored previous frame and the blurry current frame in \cite{zhou2019spatio}, misalignment also may occur due to the blurry current frame.

To overcome the problems of previous approaches, we propose a deconvolution and aggregation network (DAN) for video deblurring that utilizes both deconvolution and aggregation schemes via deep neural networks as shown in Figure~\ref{fig:overall}. In the deconvolution part, deblurring is conducted by the two-stage process with two sub-networks: the preprocessing network (PPN) and the alignment-based deconvolution network (ABDN). These networks internally solve the inverse problem of the blur model, that is why we named them as the deconvolution network, to obtain a latent frame from input blurry frames. To avoid the performance degradation due to misalignment, PPN first removes motion blurs on the input frames with non-local operations \cite{wang2018non}. ABDN then conducts frame-alignment-based deblurring to the outputs from the PPN. Because the input frames have been deblurred enough to accurately estimate object motion, the frame alignment can be performed without misalignment in ABDN.

After the deconvolution part, the frame aggregation network (FAN) performs aggregation-based deblurring by using the output frames of the ABDN. In FAN, a reliability map which represents the pixel-wise sharpness is generated for each input frame. The additional performance improvement can be achieved by merging sharp pixels of the multiple frames in accordance with reliability maps. Consequentially, a proper combination of three sub-networks (i.e. PPN, ABDN, and FAN) provides the superior performance in video deblurring by using the neighbor frames properly. Experimental results showed that the proposed method outperformed other state-of-the-art video deblurring methods in both quantitative and qualitative evaluations on the benchmark datasets \cite{su2017deep} and \cite{GOPRO_Nah_2017_CVPR}.

\section{Related Works}
\label{sec: Related Works}

\subsection{Multi-image/Video Deblurring Based on Deconvolution}

The relationship between blur frames and latent frames can be modeled by blur kernels. Many deconvolution-based deblurring methods, which remove motion blurs on the blur frame by solving the inverse problem, have been proposed \cite{kundur1996blind,shan2008high,cho2009fast,xu2010two}. In multi-image/video deblurring, inverse problems of deblurring can be solved more precisely by using additional information of multiple images.

Kim and Lee \cite{hyun2015generalized} proposed a new energy model using pixel-wise kernel estimation with optical flow for temporal coherence. In \cite{ren2017video}, a true motion trajectory is approximated for video deblurring by using the pixel-wise non-linear kernel model. In addition, deblurred images are obtained by using fast alternating minimization \cite{sroubek2011robust}, a flash gradient constraint \cite{zhuo2010robust}, a non-uniform motion blur point spread function \cite{cho2012registration}, a coupled adaptive sparse prior \cite{zhang2013multi}, and a scene model \cite{wulff2014modeling}.

\subsection{Video Deblurring Based on Aggregation}

The classical lucky imaging algorithms, such as those introduced in \cite{law2006lucky} and \cite{joshi2010lucky}, restored a high-quality image by fusing multiple low-quality images. Likewise, sharp pixels may be acquired from neighbor frames of a target frame in video deblurring. Cho et al. \cite{cho2012video} presented a video deblurring method that uses patch-based synthesis with sharpness measurement. In \cite{cho2012video}, blurry images were merged in the image domain, whereas Delbracio and Sapiro proposed video deblurring methods that aggregate multiple images in the Fourier domain \cite{delbracio2015hand,delbracio2015burst}.

\subsection{Video Deblurring Based on Deep Learning}

Recently, many video deblurring approaches based on deep learning have been proposed, and they have achieved superior results on video deblurring. Su et al. \cite{su2017deep} showed that the convolutional neural network (CNN) adequately removes motion blurs with a stacked input, which consists of a target central frame and its neighbor frames. To further improve deblurring performance on deblurring, Zhang et al. \cite{zhang2018adversarial} proposed a CNN-based method that employs generative adversarial networks (GANs) \cite{goodfellow2014generative} to impose an adversarial loss, which induces the generated outputs to look more realistic and 3D convolutions \cite{ji20123d} to capture joint spatio-temporal features. In \cite{chen2018reblur2deblur}, a deblurred image is re-blurred by the estimated pixel-wise blur kernels, and the self-supervised loss between an input blurry image and the re-blurred image helps the network to improve the deblurring performance.

Many RNN-based video deblurring approaches have also been proposed. In \cite{hyun2017online}, a feature map of the previous time step is entered into networks with a feature map of the current blurry image. The previous and current feature maps are then fused by a dynamic temporal blending network for propagation of the temporal information. In \cite{nah2019recurrent}, a hidden state is iteratively updated, which is called intra-frame iterations, to fit the propagated hidden state to the target frame. More recently, Zhou et al. \cite{zhou2019spatio} proposed a video deblurring method that utilizes a spatio-temporal filter. In \cite{zhou2019spatio}, triplet images (blurry and restored images of the previous time step, and current input blurry image) are used to capture both the motion information across frames and the blur kernel information for spatio-temporal filters.

\begin{figure*} [t]
\centering
\includegraphics[width=0.9\textwidth]{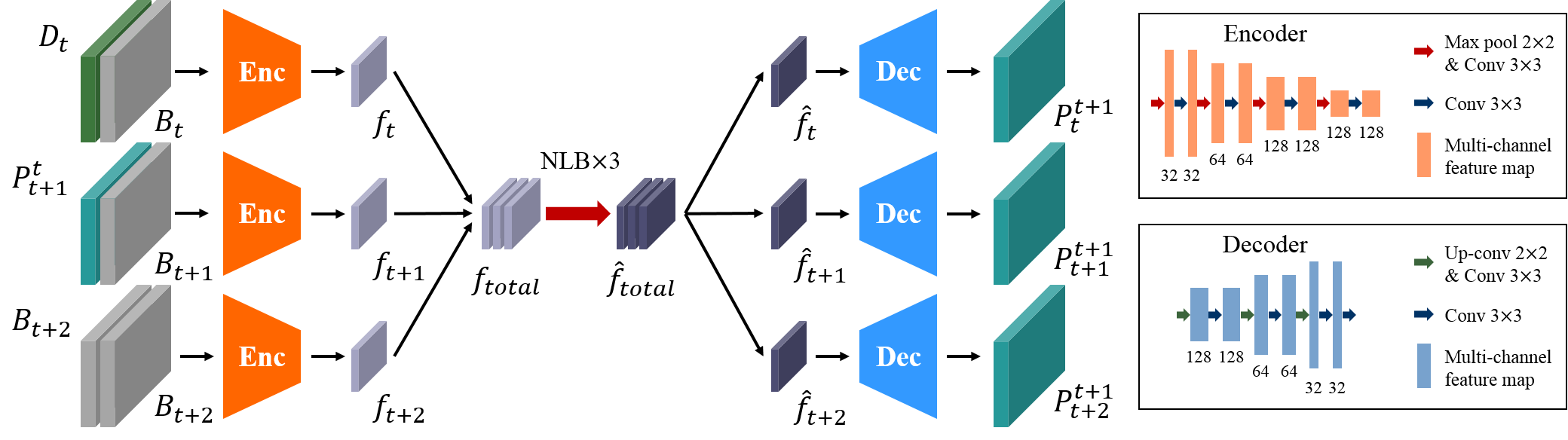}
\caption{Network architecture of the preprocessing network (PPN).}
\label{fig:PPN}
\end{figure*}

\section{The Proposed Method}
\label{sec: The Proposed Method}

\subsection{Overview}

The proposed deconvolution and aggregation network (DAN) consists of three consecutive sub-networks to obtain a latent sharp frame from input blurry frames as shown in Figure~\ref{fig:overall}. First, the blurry input frames are entered into a preprocessing network (PPN) with restored frames from the previous time step. In the PPN, the blurry frames are enhanced by referring to each other via non-local operations \cite{wang2018non}. Then, the preprocessed frames from the PPN are deblurred by an alignment-based deconvolution network (ABDN). To properly deblur the frames in the ABDN, the neighbor frames are aligned with the central frame by using optical flow. Finally, three deblurred frames, one from the ABDN and the others from the previous time steps, are merged according to reliability maps in the frame aggregation network (FAN). The aggregated frame from the FAN is the final output of the DAN. Detailed explanations of each sub-network are provided in the following sections.

\subsection{Preprocessing with Non-local Operations}

The PPN adopts the non-local neural networks proposed in \cite{wang2018non} to enhance blurry input frames. In \cite{yi2019progressive}, they showed that non-local operations could be used to reconstruct a high-resolution output from multiple low-resolution inputs. Similarly, the non-local operation can be utilized to enhance blurry pixels of the input frames as shown in Figure~\ref{fig:PPN}. If a pixel from the neighbor frame is sharp and closely related to a target pixel, a higher weight value will be assigned when the target pixel is obtained by combining.

To achieve deblurring performance and reduce the computational complexity, we adopt the recurrent manner in DAN. Consecutive blurry frames are entered into the PPN with the restored frames from the previous time step. In the PPN, the blurry frame \textit{B} and its corresponding preprocessed frame \textit{P} or deblurred frame \textit{D} (\textit{or B if restored frame is not available}) are grouped together and entered into the encoder of the PPN as shown in Figure~\ref{fig:PPN}. Then, the encoder of the PPN, $e_{PPN}$, extracts features from the grouped frames as follows:
\begin{equation}
\label{}
(f_{t},f_{t+1},f_{t+2})=e_{PPN}(B_{t}\ or\ D_{t}, B_{t+1}\ or\ P_{t+1}^{t}, B_{t+2}),
\end{equation}
where $f_{t}$ is the extracted feature map for the $t^{th}$ frame. The extracted features from each frame group are merged and processed by non-local operation as follows:
\begin{equation} \label{}
f_{total}=<f_{t},f_{t+1},f_{t+2}>,
\end{equation}
\begin{equation} \label{}
\hat{f}_{total}=NLB^{(3)}(f_{total}),
\end{equation}
where $f_{total}$ and $\hat{f}_{total}$ denote the concatenated feature maps and its output of the non-local operations, respectively. Here, $NLB^{(m)}$ denotes non-local operations with consecutive $m$ non-local blocks.

In contrast to \cite{yi2019progressive}, which directly performs the non-local operation on input frames, we perform the non-local operation on the extracted feature maps to prevent immense memory usage. The refined feature maps from the non-local operation are separated into three feature maps ($\hat{f}_{t}, \hat{f}_{t+1}, \hat{f}_{t+2}$), and the separated feature map is fed to the decoder of the PPN, $d_{PPN}$. The preprocessed frame \textit{P} is obtained for $\{t,t+1,t+2\}$ by the decoder, $d_{PPN}$, as follows:
\begin{equation} \label{}
(P_{t}^{t+1}, P_{t+1}^{t+1}, P_{t+2}^{t+1})=d_{PPN}(\hat{f}_{t}, \hat{f}_{t+1}, \hat{f}_{t+2}).
\end{equation}

Because the non-local operation directly considers relations between two frames without any frame alignment, we can effectively enhance blurry input frames, where the result is not affected by misalignment of the blurry frames. Besides, the PPN has the strength that multiple input frames are fed into the network, and the same number of frames are produced, whereas direct or slow-fusion strategies like those of \cite{su2017deep} and \cite{zhang2018adversarial} require more inputs to obtain a single output. Therefore, the PPN can adequately and instantly generate outputs without any extra inputs by adopting the non-local operation.

\subsection{Deblurring with Optical Flow-Based Alignment}

Although the blurry pixels of the input frames are enhanced by the PPN, it is difficult to eliminate all motion blurs at once. To remove motion blurs more certainly, it is necessary to solve the inverse problem of the blur model by using neighbor frames as well as the central frame. Therefore, we make the ABDN internally solve the inverse problem for aligned inputs. In ABDN, the preprocessed frames from the PPN are aligned with the central frame by using optical flow and stacked together as the input to the network. Because plenty of information of the same object is arranged in the same position, it is beneficial for the network to deblur the central frame. In addition, frame alignment is applied to the preprocessed frames from the PPN, so the alignment can be more accurate than when it is applied to the blurry input frames.

\begin{figure*} [t]
\centering
\includegraphics[width=0.9\textwidth]{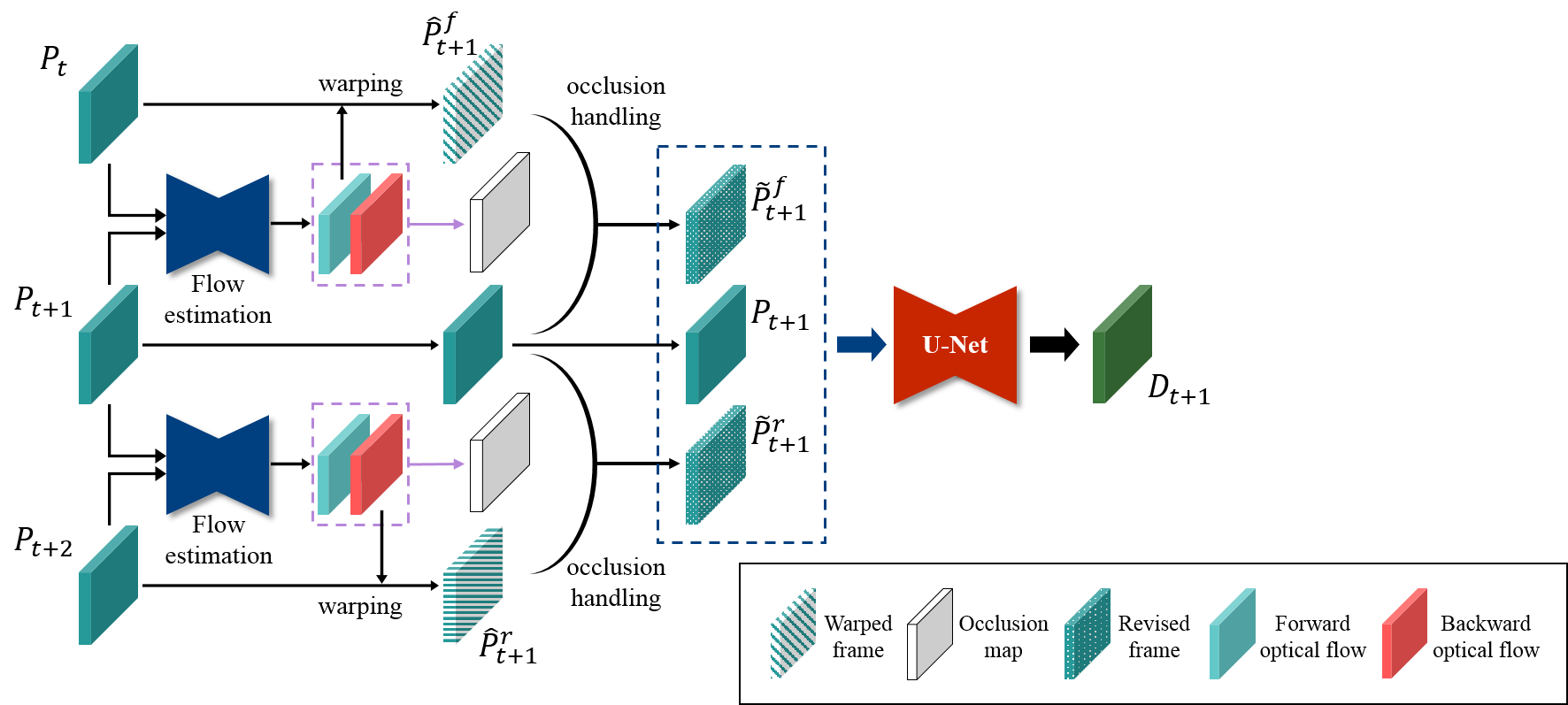}
\caption{Network architecture of the alignment-based deconvolution network (ABDN).}
\label{fig:ABDN}
\end{figure*}

As shown in Figure~\ref{fig:ABDN}, three successive preprocessed frames, $P_{t}$, $P_{t+1}$, and $P_{t+2}$ (for simplicity we dropped the time step superscript), are entered into the ABDN. Then, by using optical flow estimation algorithms like \cite{ranjan2017optical} and \cite{hui2018liteflownet}, we obtain both forward and backward optical flow maps between the adjacent frames. To align the neighbor frames with the central frame, neighbor frames of $P_{t}$ and $P_{t+2}$ are warped according to the optical flow maps, which results in the warped frames of $\hat{P}_{t+1}^{f}$ and $\hat{P}_{t+1}^{r}$, respectively.

By warping of the neighbor frames to the central frame, the warped frames can represent the central frame, but their pixel values come from the neighbor frames. When the pixels in the central frame do not exist in the neighbor frame, distortion occurs in the corresponding area of the warped frame, which is called occlusion. Because the pixels in an occlusion area in the warped frames are distinct from the pixels in the central frame, they may cause performance degradation in the deblurred image.

To prevent degradation due to occlusion, the occlusion areas of the warped frames should be detected and handled in a special way. As suggested in \cite{meister2018unflow}, an occlusion area can be detected by measuring the consistency between the forward and backward optical flow maps. Because the pixels of the central frame are not visible in the neighbor frames for occlusion areas, the difference between the forward and the backward optical flows in the occlusion region is quite large. Therefore, we consider the pixels as an occlusion region when the following inequality (\ref{eq: occ_handling}) is violated, which is introduced in \cite{meister2018unflow}:
\begin{equation}\label{eq: occ_handling}
|\mathbf{w}^{f}(\mathbf{x})+\mathbf{w}^{b}(\mathbf{x}+\mathbf{w}^{f}(\mathbf{x}))|<\alpha_{1}(|\mathbf{w}^{f}(\mathbf{x})|^{2}+|\mathbf{w}^{b}(\mathbf{x}+\mathbf{w}^{f}(\mathbf{x}))|^{2})+\alpha_{2},
\end{equation}
where $\mathbf{w}^{f}(\mathbf{x})$ and $\mathbf{w}^{b}(\mathbf{x})$ are the forward and backward optical flow maps, respectively. Here, $\alpha_{1}$ and $\alpha_{2}$ are constant values that control the inequality and were set as 0.01 and 0.5, respectively, in all our experiments. With this constraint, we can obtain an occlusion map, $Occ$, which has 0 for occluded pixels and 1 for non-occluded pixels.

The warped frames are revised to $\tilde{P}_{t+1}^{f}$ and $\tilde{P}_{t+1}^{r}$ by borrowing pixels of the central frame for the occluded pixels as follows:
\begin{equation} \label{}
\tilde{P}_{t+1}^{f}=P_{t+1}\odot(\mathbf{1}-Occ_{t})+\hat{P}_{t+1}^{f}\odot Occ_{t},
\end{equation}
\begin{equation} \label{}
\tilde{P}_{t+1}^{r}=P_{t+1}\odot (\mathbf{1}-Occ_{t+2})+\hat{P}_{t+1}^{r}\odot Occ_{t+2},
\end{equation}
where $\odot$ means pixel-wise multiplication. Then, these revised frames are stacked together with the central frame and fed into the ABDN. By using plenty of information of stacked pixels, the central frame can be effectively deblurred, and a deblurred frame, $D_{t+1}$, is obtained in the ABDN.

\begin{figure*} [t]
\centering
\includegraphics[width=0.9\textwidth]{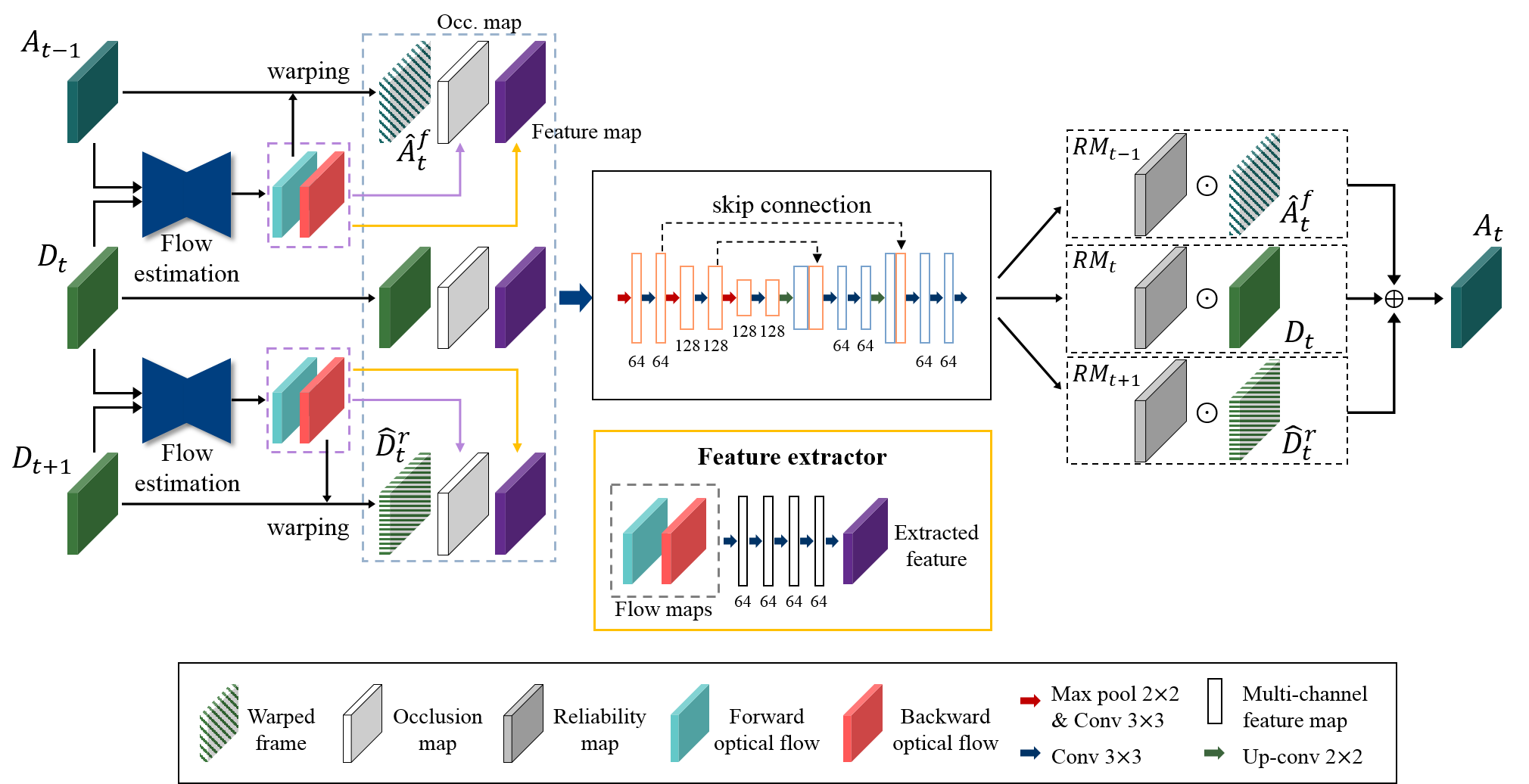}
\caption{Network architecture of the frame aggregation network (FAN).}
\label{fig:FAN}
\end{figure*}

\subsection{Frame Aggregation According to Reliability Maps}

Through the above steps, the deblurred frame $D_{t+1}$ is obtained. To achieve further improvement, we combine the consecutive deblurred frames according to the sharpness. Other aggregation-based deblurring methods, such as \cite{cho2012video} and \cite{delbracio2015hand}, utilize neighbor frames without any restoration. On the contrary, the FAN aggregates the frames which are already deblurred in the previous steps, and thus the optimal aggregation results can be accomplished.

As shown in Figure~\ref{fig:FAN}, deblurred neighbor frames are aligned with the central deblurred frame and used as the input of the FAN. The FAN then produces reliability maps (RMs), which infer the reliability of pixels in terms of sharpness of each frame. The final aggregated frame is obtained by merging the restored frames from the previous time step (warped previous output $\hat{A}_{t}^{f}$ and the deblurred frame ${D}_{t}$) and the warped deblurred frame, $\hat{D}_{t}^{r}$, according to the reliability maps as follows:
\begin{equation} \label{}
A_{t}=\hat{A}_{t}^{f}\odot RM_{t-1}+D_{t}\odot RM_{t}+\hat{D}_{t}^{r}\odot RM_{t+1},
\end{equation}
where $A_{t}$ is the final aggregated frame obtained by combining the deblurred frames. To handle the occluded pixels, the estimated occlusion maps are also fed into the FAN. For each input frame, features from the estimated optical flow maps are extracted and utilized when reliability maps are produced.

\section{Implementations and Training}
\label{sec: Implementations and Training}

\subsection{Dataset}

To train our network, we employed a video deblurring dataset from \cite{su2017deep}, which contains 71 videos, each of which is 3-5 seconds long. In \cite{su2017deep}, real-world videos were captured at a very high frame rate, and their corresponding blurry videos were synthetically generated by accumulating several consecutive frames. This allowed us to obtain pairs of blurry/sharp patches to train the network for deblurring.

As \cite{su2017deep} did, we split the dataset into 61 videos as a training set and 10 videos as a test set. To train the network adequately with the limited training set, we adopted several data augmentation techniques for the training set. We first randomly chose twenty consecutive frames from a video to compose a training video sequence for our recurrent-based network. Then, we performed data augmentation for each training sequence, such as random cropping to 256$\times$256 patches, random horizontal or vertical flipping, color jittering and adding Gaussian random noise of $N(0,0.01)$ to the images whose range is [0,1].

\subsection{Network Training}

With the training dataset from \cite{su2017deep}, we could train the network with pairs of blurry/sharp patches. In our approach, three successive networks of PPN, ABDN, and FAN are utilized to produce the final sharp frame. There are several intermediate outputs, such as the preprocessed frame \textit{P} and the deblurred frame \textit{D} as well as the final aggregated frame \textit{A}. To train our entire network effectively, we applied the mean squared error (MSE) loss function to each intermediate output as well as the final output. In the PPN, the network was trained to produce the preprocessed frame \textit{P} as close to its corresponding ground-truth sharp frame \textit{S} as possible with the MSE loss function as follows:
\begin{equation} \label{}
L_{PPN}=\mathrm{MSE}(P),
\end{equation}
where $\mathrm{MSE}(\cdot)$ denotes the mean squared error with respect to the ground-truth frame \textit{S}. Likewise, the MSE loss function was also applied to the deblurred frame \textit{D} and the final aggregated frame \textit{A} from the ABDN and FAN, respectively, as follows:
\begin{equation} \label{}
L_{ABDN}=\mathrm{MSE}(D),
\end{equation}
\begin{equation} \label{}
L_{FAN}=\mathrm{MSE}(A).
\end{equation}
To train our entire network, the sum of all loss terms was used as the total loss function as follows:
\begin{equation} \label{}
L_{total}=L_{PPN}+L_{ABDN}+L_{FAN}.
\end{equation}

\subsection{Implementation Details}

When our network was trained, patches of 256$\times$256 pixels were used with a batch size of 5. We optimized our network with an Adam optimizer \cite{kingma2014adam} with $\beta_{1}=0.9$, $\beta_{2}=0.999$ and the initial learning rate of 0.0001, which was multiplied by 0.1 for every 400k iterations. For all networks, 3$\times$3 convolutional layers were used with ReLU activation function \cite{krizhevsky2012imagenet}. In the ABDN, U-Net \cite{ronneberger2015u} was employed to deblur the preprocessed frames. Our method was implemented in the PyTorch framework \cite{paszke2017automatic}, and an Intel Xeon E5 CPU and an NVIDIA Titan Xp GPU were utilized to train the network; it took about 8 days to converge.

\section{Experimental Results}
\label{sec: Experimental Results}

\subsection{Quantitative Evaluations}

In our experiments, the proposed method was compared with several state-of-the-art video deblurring methods, such as online video deblurring (OVD) \cite{hyun2017online}, DeBlurNet (DBN) \cite{su2017deep}, video restoration framework with enhanced deformable convolution (EDVR) \cite{41wang2019edvr} and spatio-temporal filter adaptive network (STFAN) \cite{zhou2019spatio}. The proposed method and other methods (i.e. OVD \cite{hyun2017online}, DBN \cite{su2017deep}, and STFAN \cite{zhou2019spatio}) were trained with the same training set from \cite{su2017deep}, and EDVR was trained with REDS \cite{nah2019REDS} dataset.

To quantitatively confirm the video deblurring performance of the methods, we evaluated the peak signal-to-noise ratio (PSNR) of the output deblurred frame from the ground-truth sharp frame on test datasets from both Adobe240 \cite{su2017deep} and GOPRO \cite{GOPRO_Nah_2017_CVPR}. In Table~\ref{table:results}, the total average PSNR values on the test datasets are presented. As seen in Table~\ref{table:results}, we confirmed that our approach achieved the best video deblurring performance on both test datasets.

\begin{table*}[t]
\centering
\caption{Quantitative comparison with other video deblurring methods in terms of PSNR on public datasets.}
\vspace{0.2cm}
\label{table:results}
\begin{tabular}{c|ccccc}
\hline
Method  & OVD \cite{hyun2017online}    & DBN \cite{su2017deep}    & EDVR \cite{41wang2019edvr}   & STFAN \cite{zhou2019spatio}  & Proposed        \\ \hline
Adobe240 \cite{su2017deep} & 29.885 & 30.101 & 28.479 & 31.251 & \textbf{31.618} \\
GOPRO \cite{GOPRO_Nah_2017_CVPR} & 30.809 & 30.588      & 29.326      & 32.560 & \textbf{33.003} \\ \hline
\end{tabular}
\end{table*}

\subsection{Qualitative Evaluations}

\begin{figure*} [!t]
\centering
\includegraphics[width=0.9\textwidth]{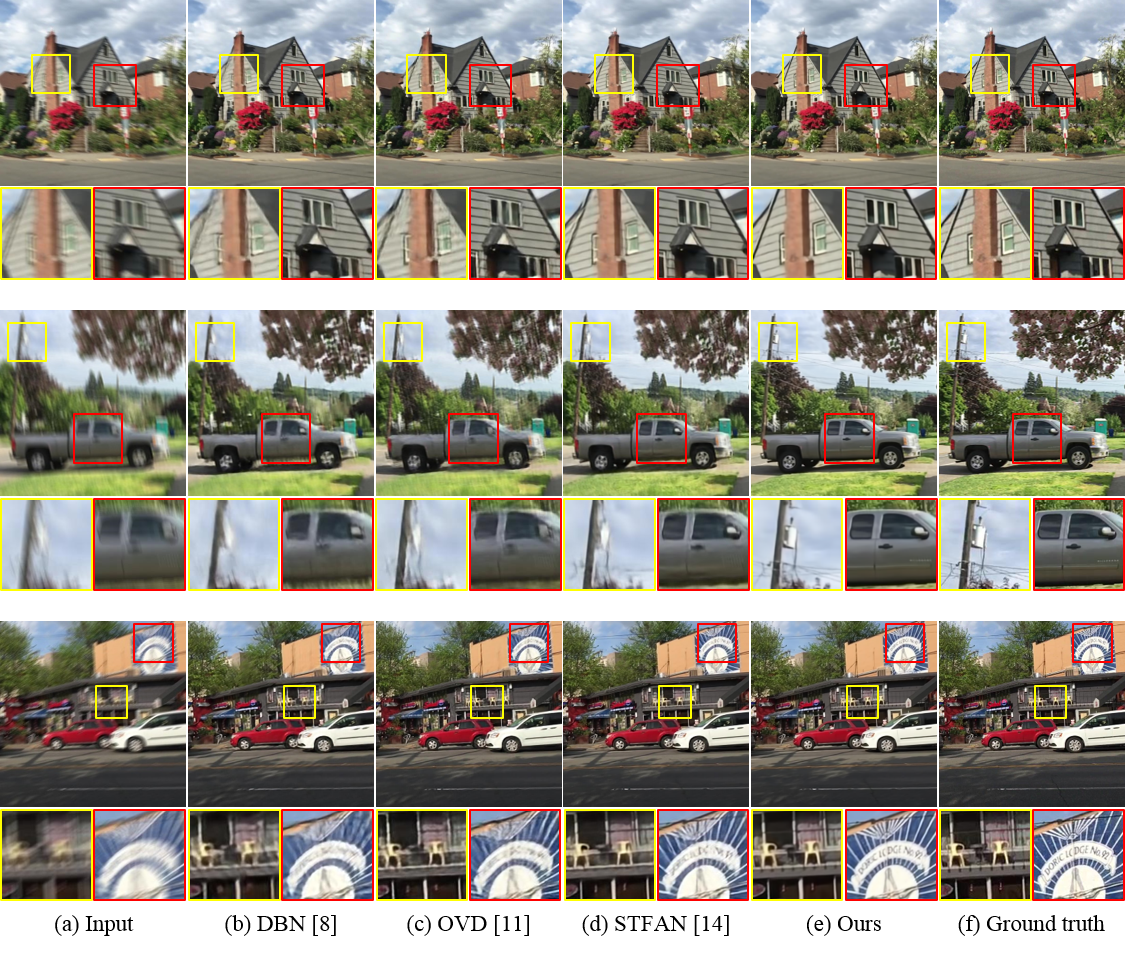}
\caption{Qualitative comparison of various video deblurring methods on the test dataset from \cite{su2017deep}. The proposed method produces sharper images especially for the yellow and red boxes.}
\label{fig:results1}
\end{figure*}

\begin{figure*} [!t]
\centering
\includegraphics[width=0.8\textwidth]{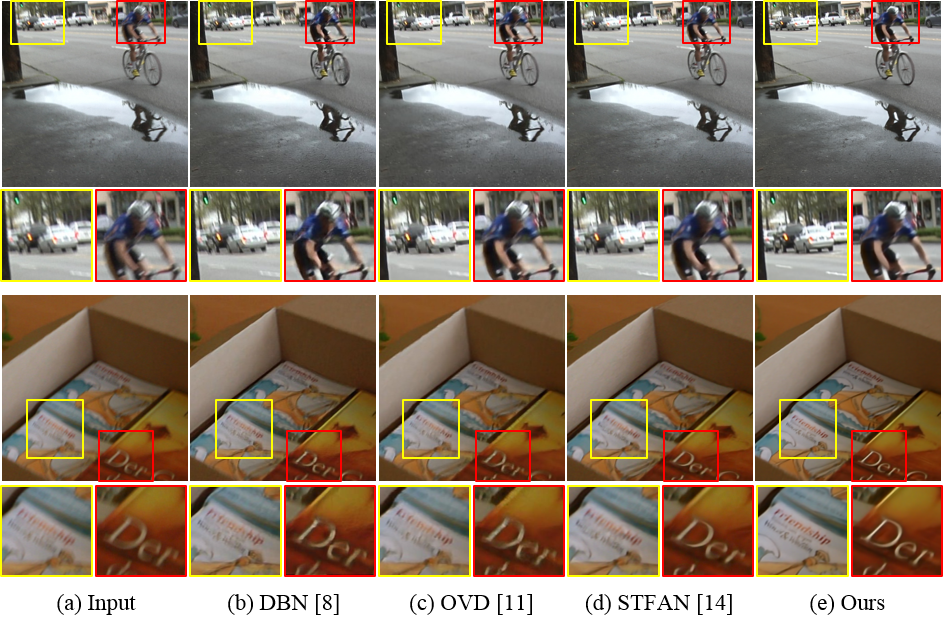}
\caption{Qualitative comparison of various video deblurring methods on the real blurry dataset from \cite{cho2012video}. The proposed method produces sharper images especially for the yellow and red boxes.}
\label{fig:results2}
\end{figure*}

To demonstrate the superiority of our approach for video deblurring, examples of deblurred frames produced by the proposed method and other methods are shown in Figures~\ref{fig:results1} and \ref{fig:results2}. These figures confirm that our approach generates deblurred frames well in comparison to the other methods on both the test set \cite{su2017deep} and real blurry images from \cite{cho2012video}. The final aggregated frames of our approach contain more detailed regions than the other approaches. In particular, even if the input frame is quite blurry, our approach removes motion blurs well while the other methods produce blurry outputs. Because our approach adopts frame alignment on the preprocessed frames, neighbor frames can be actively utilized to remove motion blurs even if the input frame is quite blurry.

\subsection{Network Size and Processing Time}

The network size and processing time of each video deblurring method are shown in Table~\ref{table:size}. In our approach, the state-of-the-art optical flow estimation methods \cite{ranjan2017optical},\cite{hui2018liteflownet} are used to align the frames in the ABDN and FAN. In our experiments, using LiteFlowNet \cite{hui2018liteflownet} produced slightly better results than using SPyNet \cite{ranjan2017optical}. However, SPyNet, which contains 1.2M parameters, is much smaller than LiteFlowNet, which contains 5.37M parameters. In addition, the network size of our approach can be adjusted by changing the size of the ABDN. While the ABDN of our full version, which utilizes U-Net \cite{ronneberger2015u}, has 13.39M parameters, only 1.05M parameters are needed for the ABDN of our \textit{light} version, which also outperforms the other video deblurring methods.

The execution time of each method is also shown in Table~\ref{table:size}. As seen in the table, our approach takes a longer time than OVD and STFAN. Because DBN has no acceleration techniques, it takes a much longer time than the other methods. The processing time of our approach may be further improved by implementing with CUDA.

\subsection{Effectiveness of Each Module}

In our approach, three successive modules are utilized to produce the deblurred frames. In this section, we analyze the effectiveness of each network module. The average PSNR values of output frames from each module are shown in Table~\ref{table:module}. These results confirm that the video deblurring performance is improved through each module. 
According to Table~\ref{table:module}, the performance of PPN with the non-local operation is higher than that of PPN without the non-local operation. We can confirm that the non-local operation helps to improve the deblurring performance in PPN. In addition, the average PSNR value is significantly increased when the ABDN is adopted after the PPN. This means that the ABDN effectively enhances the preprocessed frames that are obtained from the PPN.

\begin{table*}[t]
\centering
\caption{Network size, processing time, and performance of each video deblurring method. \textit{Light} and \textit{full} versions of our approach have 1.05M and 13.39M parameters in ABDN, respectively. Besides, our approach shows different deblurring results according to the optical flow estimation methods of \cite{ranjan2017optical} and \cite{hui2018liteflownet}.}
\vspace{0.2cm}
\label{table:size}
\newcolumntype{a}[1]{>{\raggedright\arraybackslash}p{#1}}
\newcolumntype{b}[1]{>{\centering\arraybackslash}p{#1}}
\newcolumntype{c}[1]{>{\centering\arraybackslash}p{#1}}
\newcolumntype{i}[1]{>{\centering\arraybackslash}p{#1}}
\newcolumntype{d}[1]{>{\centering\arraybackslash}p{#1}}
\newcolumntype{e}[1]{>{\centering\arraybackslash}p{#1}}
\newcolumntype{f}[1]{>{\centering\arraybackslash}p{#1}}
\newcolumntype{g}[1]{>{\centering\arraybackslash}p{#1}}
\newcolumntype{h}[1]{>{\centering\arraybackslash}p{#1}}
\begin{tabular}{a{2.1cm}|b{0.83cm} c{0.83cm} i{0.83cm} d{1.13cm} e{1.13cm} f{1.13cm} g{1.03cm} h{1.03cm}}
\hline
\shortstack{Method \\ \  } & \shortstack{OVD \\ \cite{hyun2017online} } & \shortstack{DBN \\ \cite{su2017deep} } & \shortstack{EDVR \\ \cite{41wang2019edvr} } & \shortstack{STFAN \\ \cite{zhou2019spatio} } & \shortstack{Ours \textit{light}\\ w/ \cite{ranjan2017optical}} & \shortstack{Ours \textit{light}\\ w/ \cite{hui2018liteflownet}} & \shortstack{Ours \\ w/ \cite{ranjan2017optical}} & \shortstack{Ours \\ w/ \cite{hui2018liteflownet}} \\ \hline
Params (M)              & 0.92                 & 16.67                & 23.60 & 5.37                   & 5.75                 & 9.93                 & 17.62                 & 21.80                 \\
Time (s/frame)        & 0.13                 & 1.64             & 0.21    & 0.15                   & 1.10                 & 1.19                 & 1.12                 & 1.21                 \\
Avg. PSNR               & 29.89                & 30.10         & 28.48       & 31.25                  & 31.34                & 31.41                & 31.56                & 31.62                \\ \hline
\end{tabular}
\end{table*}

\begin{table}[t]
\centering
\caption{Video deblurring performance in terms of average PSNR for various combinations of the proposed modules.}
\label{table:module}
\begin{tabular}{cc|cc}
\hline
Component     & Avg. PSNR & Component    & Avg. PSNR \\ \hline
-           & -    & PPN (w/o NLB)         & 30.187    \\
ABDN           & 30.355    & PPN (w/ NLB)         & 30.299    \\
ABDN+ABDN       & 30.472    & PPN+ABDN     & 31.228    \\
ABDN+ABDN+FAN   & 31.052    & PPN+ABDN+FAN & \textbf{31.618}    \\ \hline
\end{tabular}
\end{table}

To verify the effect of frame alignment on the deblurring process, we tried to deblur the blurry frames by using the ABDN twice. As shown in Table~\ref{table:module}, the performance of the combination of ABDN+ABDN is much worse than the combination of PPN+ABDN.  This is because the deblurred frames from the ABDN contain artifacts due to misalignment on blurry frames, and the artifacts may not be corrected by sequential adoption of ABDN. On the other hand, our approach achieves good performance due to frame alignment on the preprocessed frames from the PPN.

\subsection{Effectiveness of the Occlusion Maps}

\begin{table}[t]
\centering
\caption{Video deblurring performance in terms of PSNR with and without occlusion maps in ABDN and FAN.}
\label{table:occ}
\begin{tabular}{l|c}
\hline
Component          & Avg. PSNR \\ \hline
ABDN w/o occ. maps & 31.10     \\
ABDN w occ. maps   & 31.23     \\
FAN w/o occ. maps  & 31.57     \\
FAN w occ. maps    & 31.62     \\ \hline
\end{tabular}
\end{table}

\begin{figure*} [!t]
\centering
\vspace{0.4cm}
\includegraphics[width=0.9\textwidth]{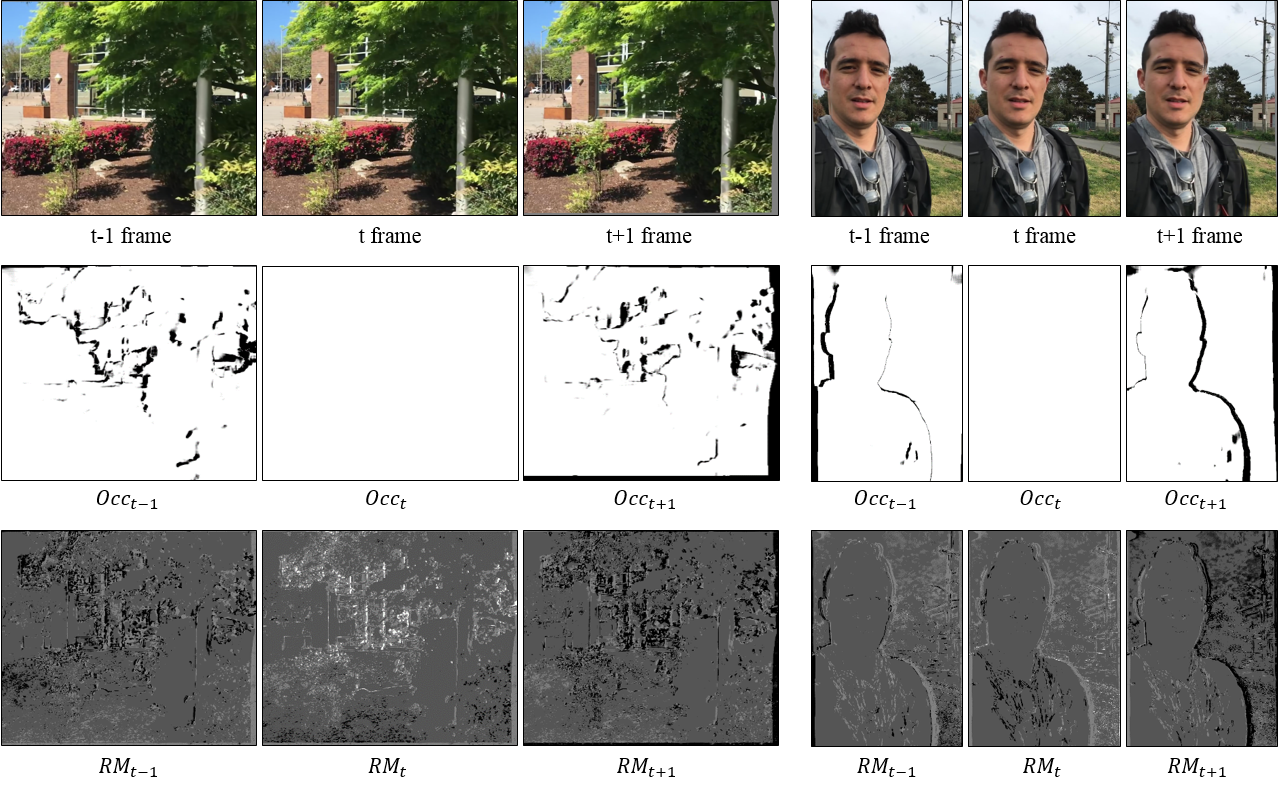}
\caption{Consecutive input frames of FAN and its corresponding occlusion maps and reliability maps.}
\label{fig:occ}
\end{figure*}

In our approach, we specially treat the occluded pixels to prevent degradation due to occlusion regions. In this section, we analyze the effectiveness of occlusion maps in the ABDN and FAN. The performance of the ABDN and FAN with and without occlusion maps is shown in Table~\ref{table:occ}. Both the ABDN and FAN showed improved performance when occlusion maps were utilized. In the ABDN, the occlusion map helps to improve the deblurring performance for occlusion regions. Moreover, occlusion maps help the FAN to generate more accurate reliability maps. Examples of the relationship between occlusion maps and reliability maps are shown in Figure~\ref{fig:occ}. As seen in the figure, we confirmed that reliability maps often have low values for occluded pixels, which can be regarded as the influence of the occlusion maps.

\section{Conclusion}
\label{sec: Conclusion}

In this paper, we proposed a new video deblurring method that employed both deconvolution and aggregation techniques via deep neural networks. Because we enhanced the blurry inputs by adopting non-local operations before alignment and deblurring, we could properly obtain deblurred frames without undesired artifacts due to misalignment. Moreover, further improvement in deblurring results could be achieved by merging multiple deblurred frames. We showed that our approach outperformed other state-of-the-art video deblurring methods through both quantitative and qualitative evaluations, which demonstrated that our approach employs the information of neighbor frames more properly.

\bibliographystyle{unsrt}  
\bibliography{references}

\end{document}